\documentclass[runningheads]{llncs}
\usepackage{graphicx}

\usepackage{caption}
\usepackage{microtype}
\usepackage{booktabs}
\usepackage{multirow} 
\newcommand{\twoLines}[2]{$\text{#1}\atop\text{#2}$}
\usepackage{listings}
\usepackage{ifxetex}
\usepackage{fancyhdr}
\usepackage{hyperref}
\usepackage{graphicx}
\usepackage{wrapfig}
\usepackage{amsmath}
\usepackage{amsfonts}
\usepackage{amssymb}
\usepackage{listings}
\usepackage{setspace}
\usepackage{wrapfig}
\usepackage{tablefootnote}
\usepackage{multirow}
\usepackage{booktabs}
\usepackage[section]{placeins}
\usepackage{color}
\usepackage{amsfonts}
\usepackage{ulem}
\usepackage{algorithm}
\usepackage{algorithmicx}
\usepackage{algpseudocode}
\usepackage{graphicx}
\usepackage{float}
\usepackage{dsfont}
\usepackage{cleveref}

\newcommand{\mat}[1]{\mathbf{#1}}
\newcommand{\set}[1]{\mathcal{#1}}

\DeclareMathOperator*{\loss}{{\ell}}

\newcommand{\N}{\ensuremath{\mathcal{N}}} 

\newcommand{\E}{\mathbb{E}}
\newcommand{\RN}{\mathbb{R}}

\newcommand{\Indicator}{\mathds{1}}				

\DeclareMathOperator*{\regularization}{\ensuremath{\theta}}
\DeclareMathOperator*{\densitythreshold}{\ensuremath{\delta}}
\newcommand{\density}{\ensuremath{p}}
\newcommand{\densityestimator}{\ensuremath{\hat{p}}}
\newcommand{\densityestimatorkde}{\densityestimator_{\text{KDE}}}
\newcommand{\densityestimatorgmm}{\densityestimator_{\text{GMM}}}
\newcommand{\generatingprocess}{\ensuremath{\Psi}}
\newcommand{\x}{\ensuremath{\vec{x}}}
\newcommand{\z}{\ensuremath{\vec{z}}}
\newcommand{\y}{\ensuremath{y}}
\newcommand{\setx}{\ensuremath{\set{X}}}
\newcommand{\sety}{\ensuremath{\set{Y}}}
\newcommand{\xcf}{\ensuremath{\vec{x}'}}
\newcommand{\ycf}{\ensuremath{y^{c}}}
\newcommand{\classifier}{\ensuremath{h}}
\newcommand{\classifierset}{\ensuremath{\set{H}}}

\DeclareMathOperator*{\distmat}{{\mat{\Omega}}}

\newboolean{arxiv}
\setboolean{arxiv}{true} 

\begin{document}
\title{Convex Density Constraints for Computing Plausible Counterfactual Explanations\thanks{We gratefully acknowledge funding from the VW-Foundation for the project \textit{IMPACT} funded in the frame of the funding line \textit{AI and its Implications for Future Society}.}}

\titlerunning{Computation of Plausible Counterfactual Explanations}
%
\author{Andr\'e Artelt\inst{1} \and Barbara Hammer\inst{1}}
\authorrunning{A. Artelt and B. Hammer}
%
\institute{CITEC - Cognitive Interaction Technology\\Bielefeld University, 33619 Bielefeld, Germany\\
\email{\{aartelt,bhammer\}@techfak.uni-bielefeld.de}}
\maketitle              
\begin{abstract}
The increasing deployment of machine learning as well as legal regulations such as EU's GDPR cause a need for user-friendly explanations of decisions proposed by machine learning models.
Counterfactual explanations are considered as one of the most popular techniques to explain a specific decision of a model.
While the computation of ''arbitrary'' counterfactual explanations is well studied, it is still an open research problem how to efficiently compute plausible and feasible counterfactual explanations.
We build upon recent work and propose and study a formal definition of plausible counterfactual explanations. In particular, we investigate how to use density estimators for enforcing plausibility and feasibility of counterfactual explanations. For the purpose of efficient computations, we propose convex density constraints that ensure that the resulting counterfactual is located in a region of the data space of high density.

\keywords{XAI \and Counterfactual Explanations \and Transparency \& Interpretability.}
\end{abstract}
\section{Introduction}
As research on machine learning (ML) is making more and more progress and ML models constitute state-of-the-art approaches in domains such as machine translation, image and text classification, we observe an increased deployment of ML technology in practice~~\cite{Goel2016PrecinctOP,creditriskml,creditscoresunfair}. 
At the same time, ML models are vulnerable to 
unexpected behavior such as adversarial attacks~\cite{adversarialattacks} and
behavior which is regarded as unfair by humans~\cite{fairml},
hence a large amount of the decision making process offered by ML is not fully understood by humans.
As a consequence of this fact and due to legal regulations like EU's GDPR~\cite{gdpr}, transparency and interpretability of ML models becomes more and more relevant.  Therefore, there is a need for tools that make ML models transparent in the sense that we can explain the decision making process of a model. Accordingly, we observe an increase of research in the area of explainable AI (XAI)~\cite{explainingexplanations,explainingblackboxmodelssurvey,explainableartificialintelligence,surveyxai}.

Over time, researchers developed a diverse set of methods for explaining ML models~\cite{explainingblackboxmodelssurvey,molnar2019}:
Model-agnostic methods~\cite{explainingblackboxmodelssurvey,modelagnosticinterpretability}  are not tailored to a particular model or representation, hence they are (in theory) applicable to any different types of ML models; in the extreme "truly" model-agnostic methods do not need access to the training data or model internals but they regard the model as a black-box.
There exists a variety of different model-agnostic approaches, including
feature interaction methods~\cite{featureinteraction}, feature importance methods~\cite{featureimportance}, partial dependency plots~\cite{partialdependenceplots} and local methods that approximates the model locally by an explainable model~\cite{decisiontreecounterfactual,lime2016}.
This group of technologies relies on feature importance ranking or similar to express decisions of a given model.
A different class of explanations relies on examples that explain a prediction by a (set of) data points~\cite{casebasedreasoning}. Prototypes \& criticisms~\cite{prototypescriticism} and influential instances~\cite{influentialinstances} are  instances of such example-based explanations.

One popular instance of example-based explanations, often realized as black-box scheme, are counterfactual explanations~\cite{molnar2019,counterfactualwachter}. A counterfactual explanation states a change to the original input that leads to a different prediction of a given ML model. This type of explanation is considered as particularly intuitive, because it tells the user what to do in order to achieve a desired goal~\cite{molnar2019,counterfactualwachter}.
Despite the huge variety of different - equally important - types of explanations, we limit ourselves to  counterfactual explanations in this contribution.
Counterfactual
explanations can be phrased as a constrained optimization problem,
aiming for minimizing the change which results in a different output.
Depending on the specific setting,  this optimization problem is solved by either gradient-based schemes or, in particular in agnostic settings, by black-box solvers. Thereby, approaches which rely on the specific form
of the given classifier can lead to much more efficient computation schemes, as demonstrated in \cite{counterfactualslvq}.

Yet, stated in its simplest form, counterfactuals are very similar to adversarial examples, since there are no guarantees that the resulting counterfactual is plausible and feasible in the data domain.
As a consequence, the absence of such  constraints often leads to counterfactual explanations that are not plausible~\cite{contrastiveexplanations,counterfactualguidedbyprototypes,face} - an observation that we will also   confirm in this work.

In this work, we aim for an extension of counterfactual explanation schemes which restricts possible explanations to plausible regions of the data space. More specifically, we propose and study a formal definition of plausible counterfactual explanations and propose a modeling framework, which phrases such constraints in convex form, such that they can efficiently be integrated
into optimization schemes, preserving uniqueness of solutions or efficiency if this is valid for the constrained version.

\section{Definition and Related Work}
We briefly review existing work on enforcing plausibility of counterfactual explanations (Definition~\ref{def:counterfactualexplanation}).
In the context of ML models, counterfactual explanations are  formalized as follows:
\begin{definition}[Counterfactual explanation~\cite{counterfactualwachter}]\label{def:counterfactualexplanation}
Assume a prediction function $\classifier$ is given. Computing a counterfactual $\xcf \in \RN^d$ for a given input $\x \in \RN^d$ is phrased as the following optimization problem:
\begin{equation}\label{eq:counterfactualoptproblem}
\underset{\xcf \,\in\, \RN^d}{\arg\min}\; \loss\big(\classifier(\xcf), \ycf\big) + C \cdot \regularization(\xcf, \x)
\end{equation}
where $\loss(\cdot,\cdot)$ denotes a loss function, $\ycf$ the requested prediction, and $\regularization(\cdot,\cdot)$ a penalty term for deviations of $\xcf$ from the original input $\x$. $C>0$ denotes the regularization strength.
\end{definition}

Two common regularizations are the weighted Manhattan distance and the Mahalanobis distance.
The weighted Manhattan distance is defined as:
\begin{equation}\label{eq:weighted_l1}
\regularization(\xcf, \x) = \sum_j \alpha_j \cdot |(\x)_j - (\xcf)_j|
\end{equation}
where $\alpha_j > 0$ denote  feature-wise weights.
The Mahalanobis distance is defined as:
\begin{equation}\label{eq:general_l2}
\regularization(\xcf, \x) = (\x - \xcf)^\top\distmat(\x - \xcf)
\end{equation}
where $\distmat$ denotes a s.psd. matrix.

In general, $\xcf$ is arbitrary, hence possibly implausible. A variety of approaches aims for a restriction of the domain to plausible patterns only.
The authors of~\cite{face} propose to compute a path of intermediate counterfactuals that lead to the final counterfactual. The idea  is to provide the user with a set of intermediate goals that finally lead to the desired goal - it might be easier to ``go into the direction" of the final goal step by step instead of accomplishing it in a single step. In order to compute such a path of intermediate counterfactuals, different strategies for constructing a graph on the training data set are proposed - including the query point. In this graph, two samples are connected by a weighted edge if they are ``sufficiently close to each other" - e.g.\ based on density estimation. The path of intermediate counterfactuals is then computed as the shortest path between the query point and a point that has the requested label - this data point is the final counterfactual. Therefore, the final counterfactual as well as all intermediate counterfactuals are elements of the training data set, which ensures that all counterfactuals are plausible and feasible.
However, the limitation to samples from the training set can be seen as a major drawback of this method, in particular for sparsely populated data spaces.

A slightly modified version of Eq.~\eqref{eq:counterfactualoptproblem} was proposed in~\cite{counterfactualguidedbyprototypes}. The authors suggest that the original formalization Eq.~\eqref{eq:counterfactualoptproblem} does not take into account that the counterfactual should lie on the data manifold which would enforce plausibility. Therefore, they propose to add two additional terms to the original objective in Eq.~\eqref{eq:counterfactualoptproblem}, which should be simultaneously optimized:
\begin{enumerate}
\item The distance between the counterfactual $\xcf$ and the reconstructed version of it that has been computed by using a pretrained autoencoder.
\item The distance between the encoding of the counterfactual $\xcf$ and the mean encoding of training samples that belong to the requested class $\ycf$.
\end{enumerate}
The first term is supposed to ensure that the counterfactual $\xcf$ lies on the data manifold and thus is a plausible data instance. The second term is supposed to accelerate the solver for computing the solution of the final optimization problem.
We think that this is a very promising approach - However, the objective itself still behaves like ''a heuristic" because, like the original Eq.~\eqref{eq:counterfactualoptproblem}, there are no guarantees that the resulting counterfactual is plausible/feasible or even valid at all - one would have to do an extensive hyperparameter tuning of the objective. Furthermore, the need of a working autoencoder can be considered as another bottleneck because building high quality and stable autoencoders can be quite challenging if only very little data are available - in particular if the autoencoder is  modeled by deep neural networks. Lastly, due to the non-convexity of the autoencoder and the model itself, the resulting optimization problem is highly non-convex and thus difficult to solve. 

Somehow similar to~\cite{counterfactualguidedbyprototypes}, the authors of~\cite{contrastiveexplanationsganvae} propose to use GANs and VAEs for creating realistic images. Although they do not talk explicitly about counterfactuals - they want to compute contrastive explanations\footnote{A contrastive explanations states a minimal amount of (present and absent) features (including their values) that are responsible for a specific prediction. Such an explanation is computed by finding a minimal perturbation to the input that yields the same (present features) or different (absent features) prediction. In order to stay close to the data manifold - enforce that the results are plausible - they propose to use an autoencoder.}~\cite{contrastiveexplanations} which are similar to counterfactuals in the sense that in both cases we want to find a minimal change that leads to a specific prediction (although we have a second objective in constrastive explanations).

The authors of~\cite{counterfactualcomputationsurvey} propose a convex modeling framework for efficiently computing counterfactual explanations of different ML models. They propose to turn the optimization problem Eq.~\eqref{eq:counterfactualoptproblem} into a constraint optimization problem:
\begin{equation}\label{eq:cf:constraintform}
\underset{\xcf \,\in\,\RN^d}{\arg\min}\;\regularization(\xcf, \x) \quad \text{ s.t. } \classifier(\xcf) = \ycf
\end{equation}
By exploiting model specific structures, they are able to turn Eq.~\eqref{eq:cf:constraintform} into a convex program for many different ML models. The benefits of this modeling are that convex programs can be solved very efficiently~\cite{Boyd2004}, additional convex constraints can be added without changing the complexity of the problem, feasibility - does a solution (counterfactual), under a given set of constraints, exist? - can be verified easily. By adding additional constraints we can ensure that the counterfactual is plausible/feasible in the specific data domain. However, manually constructing plausibility constraints can be very time consuming and requires solid domain knowledge which might not be available.
These approaches yield promising approaches, yet their greatest disadvantage is the potentially high computational load of the induced optimization problem. Here,
we will take a different avenue by phrasing the condition of plausibility as a convex constraint.

Our contribution builds on our prior work~\cite{counterfactualcomputationsurvey}, which phrases
counterfactual computation in terms of efficient constrained optimization problems for many 
popular classifiers. Besides a formal definition of plausible counterfactuals, we propose convex density constraints that can be built from a given data set automatically and efficiently. These constraints ensure that the density of the resulting counterfactual is lower bounded by a predefined/requested threshold.
\ifthenelse{\boolean{arxiv}}{Note that all proofs and derivations can be found in the appendix~\ref{sec:appendix}.}{Due to space constraints, all proofs and derivations can be found in an extended version on arXiv\footnote{\url{TODO-LINK-TO-ARXIV-VERSION-GOES-HERE}}.} 

\section{Plausible Counterfactual Explanations}
\subsection{Computation of Plausible Counterfactual Explanations}
%
For the purpose of enforcing plausibility of counterfactuals, we propose to add a target specific density constraint to Eq.~\eqref{eq:cf:constraintform}:
\begin{subequations}\label{eq:cf:densityconstraintform}
\begin{align}
&\underset{\xcf \,\in\,\RN^d}{\arg\min}\;\regularization(\xcf, \x)\label{eq:cf:densityconstraintform:objective}\\
&\quad \text{s.t. } \classifier(\xcf) = \ycf \label{eq:cf:densityconstraintform:classifierconstraint}\\
&\quad\quad\;\; \densityestimator_{\y}(\xcf) \geq \densitythreshold \label{eq:cf:densityconstraintform:densityconstraint}
\end{align}
\end{subequations}
where $\densityestimator_{\y}(\cdot)$ denotes a class dependent density estimator.

There exists a variety of different density estimators that estimate the density based on training samples.

A kernel density estimator (KDE) is a popular choice when it comes to estimate densities from training data. A kernel density estimator is a non-parametric model and is defined as:
\begin{equation}\label{eq:kde}
\densityestimatorkde(\x) = \sum_i \alpha_i k(\x, \x_i)
\end{equation}
where $k(\cdot,\cdot)$ denotes a suitable kernel function, $\x_i$ denotes the $i$-th sample in the training data set and $\alpha_i>0$ denotes the weighting of the $i$-th sample.
However, in case of non-linear kernels (e.g. Gaussian kernel) the resulting density estimator is highly non-convex and does not
induce an efficient optimization problem.

In a Gaussian mixture model (GMM) the density is modeled as a mixture of multivariate normal distributions. The density under a GMM with $m$ components is defined as:
\begin{equation}\label{eq:gmm}
\densityestimatorgmm(\x) = \sum_{j=1}^{m} \pi_j \N(\x \mid \vec{\mu}_j,\mat{\Sigma}_j)
\end{equation}
where $\pi_j$ denotes the prior probability of the $j$-th component, $\vec{\mu}_j$ and $\mat{\Sigma}_j$ denote the mean and covariance of the $j$-th component. Although the GMM Eq.\eqref{eq:gmm} is much simpler (has fewer components/parameters) than a kernel density estimator Eq.~\eqref{eq:kde}, it still does not induce convex constraints for
 Eq.~\eqref{eq:cf:densityconstraintform:densityconstraint}.

Here we propose to approximate the density of a GMM Eq.~\eqref{eq:gmm} by a component wise maximum:
\begin{equation}\label{eq:gmm:approx}
\densityestimator(\x) = \underset{j}{\max}\Big(\densityestimator_j(\x)\Big)
\end{equation}
where
\begin{equation}\label{eq:gmm:approx:component}
\densityestimator_{j}(\x) = \pi_j \N(\x \mid \vec{\mu}_j,\mat{\Sigma}_j)
\end{equation}
By construction, the approximation Eq.~\eqref{eq:gmm:approx} is always a lower bound of the true GMM density Eq.~\eqref{eq:gmm}. More precisely, the following bound holds:
\begin{equation}\label{eq:gmm:approx:bound}
\densityestimator(\x) \leq \densityestimatorgmm(\x) \leq m\cdot\densityestimator(\x)
\end{equation}
The inequality constraint of a single component Eq.~\eqref{eq:gmm:approx:component}
\begin{equation}\label{eq:gmm:approx:component:constraint}
\densityestimator_j(\x)  = \pi_j\N(\x \mid \vec{\mu}_j,\mat{\Sigma}_j) \geq \densitythreshold
\end{equation}
can be rewritten as a convex quadratic constraint:
\begin{equation}\label{eq:gmm:approx:component:finalconstraint}
(\x - \vec{\mu}_j)^\top\mat{\Sigma}_j^{-1}(\x - \vec{\mu}_j) + c_j \leq {\densitythreshold}'
\end{equation}
where
\begin{equation}
c_j = -2\log(\pi_j) + d\log\left(2\pi\right) - \log\Big(\det(\mat{\Sigma}_j^{-1})\Big) \quad\quad {\densitythreshold}' = -2 \log(\densitythreshold)
\end{equation}
By making use of the approximation Eq.~\eqref{eq:gmm:approx}, the original constraint Eq.~\eqref{eq:cf:densityconstraintform:densityconstraint} becomes:
\begin{equation}\label{eq:gmm:approx:finalconstraint}
\underset{j}{\min}\Big((\xcf - \vec{\mu}_j)^\top\mat{\Sigma}_j^{-1}(\xcf - \vec{\mu}_j) + c_j\Big)  \leq {\densitythreshold}'
\end{equation}
Although Eq.~\eqref{eq:gmm:approx:finalconstraint} is still non-convex, we can turn it into a set of convex constraints by observing the following:

Let $\xcf_{*}$ be a solution of Eq.~\eqref{eq:cf:densityconstraintform} where we substituted Eq.~\eqref{eq:cf:densityconstraintform:densityconstraint} by Eq.~\eqref{eq:gmm:approx:finalconstraint}. Then it holds that:
\begin{equation}
\exists\, j\in\{1, \dots, m\}:\;(\xcf_{*} - \vec{\mu}_j)^\top\mat{\Sigma}_j^{-1}(\xcf_{*} - \vec{\mu}_j) + c_j \leq {\densitythreshold}'
\end{equation}
Note that there might exists more than one $j$ for which Eq.~\eqref{eq:gmm:approx:component:constraint} holds. Because we do not know for which $j$ Eq.~\eqref{eq:gmm:approx:component:constraint} holds, we simply try all possible $j\in\{1, \dots, m\}$ and select the counterfactual $\xcf$ that yields the smallest value of the objective Eq.~\eqref{eq:cf:densityconstraintform:objective} - that is the closest to the original input $\x$. Note that depending on the prediction function $\classifier(\cdot)$ it can happen that Eq.~\eqref{eq:cf:densityconstraintform} is not feasible for all $j$. Because each constraint Eq.~\eqref{eq:gmm:approx:component:constraint} can be rewritten as a convex quadratic constraint, the final optimization problem Eq.~\eqref{eq:cf:densityconstraintform} becomes convex iff the objective Eq.\eqref{eq:cf:densityconstraintform:objective} and the prediction constraint Eq.~\eqref{eq:cf:densityconstraintform:classifierconstraint} are convex. The Manhattan distance as well as the Mahalanobis distance as regularizers $\regularization(\cdot,\cdot)$ together with common ML models - like generalized linear models, linear SVM, LDA, matrix LVQ, decision tree, etc. - yield convex programs~\cite{counterfactualcomputationsurvey} that can be solved efficiently~\cite{Boyd2004}.

\subsection{A Formal Approach}
We aim for a formal description of plausible counterfactuals as modelled in Eq.~\eqref{eq:cf:densityconstraintform}.

We assume a classification setting with an underlying generating process $\generatingprocess=(\setx, \sety, \density)$ where the measurable set $\setx$ denotes the data domain, the discrete and finite set $\sety$ denotes the set of possible labels and $\density: \setx \times \sety\mapsto \RN_{+}$ denotes the joint density - we assume that $\{\x\in\setx\mid\density(\x,\y) \geq \densitythreshold\}$ is closed for all $\densitythreshold>0,\y\in\sety$. Furthermore, let $\regularization:\setx\times\setx\mapsto \RN_{+}$ be a distance metric on $\setx$. Following Eq.~\eqref{eq:cf:densityconstraintform}, we propose to define a \textit{plausible counterfactual} according to Definition~\ref{def:plausiblecounterfactual}.
\begin{definition}[$\densitythreshold$-plausible counterfactual]\label{def:plausiblecounterfactual}
Let $\classifier:\setx  \mapsto \sety$ be a classifier. We call a counterfactual explanation $(\xcf, \ycf)$ of a particular sample $\x\in\setx$ $\densitythreshold$-plausible iff the following holds:
\begin{equation}\label{eq:deltaplausiblecounterfactual:optprob}
\xcf = \underset{\xcf \,\in\,\setx}{\arg\min}\;\regularization(\xcf, \x) \quad \text{s.t. } \classifier(\xcf) = \ycf \; \land \; \density(\xcf,\ycf) \geq \densitythreshold
\end{equation}
\end{definition}
where $\densitythreshold > 0$ denotes a minimum density at which we consider a sample plausible. Note that we state the definition of an $\densitythreshold$-plausible counterfactual as an optimization problem Eq.\eqref{eq:deltaplausiblecounterfactual:optprob} which makes the definition particular appealing from a practical perspective.

Next, in Theorem~\ref{theorem:modelfreecounterfactuals} we state under what conditions $\densitythreshold$-plausible counterfactuals do not depend on the classifier.
\begin{theorem}[Model free $\densitythreshold$-plausible counterfactuals under zero risk classifiers]\label{theorem:modelfreecounterfactuals}
Let $\classifierset$ be the set of all classifiers $\classifier:\setx\mapsto\sety$ that have zero risk on the generating process $\generatingprocess$ - that is: $\classifier \in \classifierset \Leftrightarrow \underset{\x, \y \sim \density}{\E}[\Indicator\left(\classifier(\x) \neq \y\right)] = 0$. Then the following holds $\forall\, \classifier\in\classifierset, (\x,\ycf)\in\setx\times\sety\setminus\{\y\}$:
\begin{equation}
\begin{split}
& \underset{\xcf \,\in\,\setx}{\arg\min}\;\regularization(\xcf, \x) \; \text{s.t. } \classifier(\xcf) = \ycf \, \land \, \density(\xcf,\ycf) \geq \densitythreshold \\
& \Leftrightarrow \underset{\xcf \,\in\,\setx}{\arg\min}\;\regularization(\xcf, \x) \; \text{s.t. } \density(\xcf,\ycf) \geq \densitythreshold
\end{split}
\end{equation}
\end{theorem}
Note that Theorem~\ref{theorem:modelfreecounterfactuals} states that in the case of perfect classifiers, $\densitythreshold$-plausible counterfactuals become independent from the specific classifiers - thus we can compute the $\densitythreshold$-plausible counterfactuals solely in the data domain without taking the classifiers into account.

However, in practice we usually do not have a perfect classifier because either the class wise densities are overlapping or the classifier itself can not model a zero risk decision boundary. Therefore, we state a weaker version of Theorem~\ref{theorem:modelfreecounterfactuals} in Theorem ~\ref{theorem:modelfreecounterfactuals:localversion}, in which we assume that a classifiers $\classifier$ is locally $\densitythreshold$-sufficient perfect at a sample $(\x, \y)$ (Definition~\ref{def:locallyperfectclassifier}) - 
that is: the classifier $\classifier$ classifies the sample $\x$ as $\y$, which is consistent with the ground truth induced by the generating process $\generatingprocess$, and the decision boundary does not "cut to deep" into the closest parts of high density regions of the other classes.
\begin{definition}[Locally $\densitythreshold$-sufficient perfect classifier]\label{def:locallyperfectclassifier}
Let $\classifier:\setx\mapsto\sety$ be a classifier and denote the set of all $\x\in\setx$ that have a class dependent density of at least $\densitythreshold$ by $\setx_{\densitythreshold}(\ycf)$ - that is: $\setx_{\densitythreshold}(\y)=\{\x\in\setx \mid \density(\x,\y)\geq\densitythreshold\}$. We call $\classifier$ locally $\densitythreshold$-sufficient perfect at a sample $(\x, \y) \in \setx\times\sety$ iff the following holds:
\begin{equation}
\classifier(\x) = \underset{\y_i\, \in\, \sety}{\arg\max}\,p(\x, \y_i)=\y \;\land\; \classifier(\x_{*}) = \ycf \;\;\forall\, \ycf \in \sety\setminus\{\y\},\; \x_{*}= \underset{\z\,\in\,\setx_{\densitythreshold}(\ycf)}{\arg\min}\regularization(\z, \x)
\end{equation}
\end{definition}
\begin{theorem}[Model free $\densitythreshold$-plausible counterfactual under locally\\ $\densitythreshold$-sufficient perfect classifiers]\label{theorem:modelfreecounterfactuals:localversion}
Let $\classifierset(\x,\y)$ be the set of locally $\densitythreshold$-sufficient perfect classifiers (Definition~\ref{def:locallyperfectclassifier}) at a sample $(\x,\y) \in \setx\times\sety$. Then the following holds $\forall\, \classifier\in\classifierset(\x,\y), (\x,\ycf)\in\setx\times\sety\setminus\{\y\}$:
\begin{equation}
\begin{split}
& \underset{\xcf \,\in\,\setx}{\arg\min}\;\regularization(\xcf, \x) \; \text{s.t. } \classifier(\xcf) = \ycf \, \land \, \density(\xcf,\ycf) \geq \densitythreshold \\
& \Leftrightarrow \underset{\xcf \,\in\,\setx}{\arg\min}\;\regularization(\xcf, \x) \; \text{s.t. } \density(\xcf,\ycf) \geq \densitythreshold
\end{split}
\end{equation}
\end{theorem}
Note that Theorem~\ref{theorem:modelfreecounterfactuals:localversion} states that for a set of classifiers that are locally $\densitythreshold$-sufficient perfect  at a sample $(\x, \y) \in \setx\times\sety$ (Definition~\ref{def:locallyperfectclassifier}), the $\densitythreshold$-plausible counterfactuals of this particular sample $\x$ are exactly the same for all classifiers in this set. Because we only assume locally $\densitythreshold$-sufficient perfectness of the classifier, Theorem~\ref{theorem:modelfreecounterfactuals:localversion} is very appealing for practice when we actually have to compute a counterfactual explanation of a particular sample under a particular model - the theorem tells us when we can drop the classification constraint and thus simplify the optimization problem Eq.~\eqref{eq:deltaplausiblecounterfactual:optprob}.

In practice, when the true density (or a density estimation) is not available, one could try to check for locally $\densitythreshold$-sufficient perfectness at a given sample $\x$ by checking if the "closest" training samples (incl. samples from different classes) around $\x$ are classified correctly.

\section{Experiments}
We perform experiments on several data sets\footnote{Our source code is available on GitHub - \url{https://github.com/andreArtelt/ConvexDensityConstraintsForPlausibleCounterfactuals}} for empirically evaluating our proposed density constraints Eq.~\eqref{eq:gmm:approx:finalconstraint}. We use the "Breast Cancer Wisconsin (Diagnostic) Data Set"~\cite{breastcancer}, the "Iris Plants Data Set"~\cite{irisdata}, the "Wine Data Set"~\cite{winedata}, the "Boston Housing Data Set"~\cite{bostonhousing}\footnote{We turn it into a binary classification problem by setting the target to $1$ if the price is greater or equal to 20k\$.} and the "Optical Recognition of Handwritten Digits Data Set"~\cite{ocr}.
We repeat the following procedure in a 4-fold cross validation:
First, we fit class dependent kernel density estimators (we use the Gaussian kernel) and a GMM to the training data set - where we use a 5-fold cross validation grid search for hyperparameter tuning.
Next, we fit a classifier (either a softmax regression or decision tree)\footnote{An implementation of the experiments including other models like LDA, linear SVM, matrix LVQ, etc. is available online.} to the training data set.
After this, for each sample in the test set, we compute two counterfactuals (both with the same but random target class) - one counterfactual without any additional density/plausibility constraints and another counterfactual with our proposed density constraint Eq.~\eqref{eq:gmm:approx:component:finalconstraint}. We set the density threshold $\densitythreshold$ from Eq.~\eqref{eq:gmm:approx:component:constraint} to the median density Eq.~\eqref{eq:gmm:approx} of the training samples under the approximated GMM of the target class $\ycf$. To enforce sparsity, both counterfactuals are computed under the Manhattan distance as a regularizer $\regularization(\cdot,\cdot)$.
Finally, we compute the Manhattan distance to the original sample and the log-density of both counterfactuals under the kernel density estimator. We use the kernel density estimator instead of the GMM because our proposed density constraint is an approximation of the GMM which itself can be interpreted as an approximation of the kernel density estimator.
In order to increase the accuracy of the classifiers and density estimators, we apply a PCA to the breast cancer data set (5 components), the house prices data set (10 components), the wine data set (8 components) and the digits data set (40 components). Since the PCA transformation is affine, it can be easily integrated into our convex programs - so that we can still compute counterfactuals in the original space.

The results of the experiments are listed in Table~\ref{tab:experiments:results}.
\begin{table}[tb]
\caption{Median log-density (under the KDE) and median Manhattan distance to the original sample of the  computed counterfactuals - with vs. without density constraints. Best values are \textbf{highlighted} - larger densities and smaller distances are better.}\label{tab:experiments:results}
\begin{tabular}{|c|c||c|c||c|c|}
\hline
&&  \multicolumn{2}{|c||}{Without density constraints} & \multicolumn{2}{|c|}{With density constraints}\\
\cline{2-6}
& Data set & Density & Distance & Density & Distance \\
\hline\hline
\multirow{5}{*}{\rotatebox[origin=l]{90}{\twoLines{Softmax}{regresssion}}}
& Iris &  -34.55 & \textbf{1.80} & \textbf{-0.75} & 4.06 \\
& Digits & -164.03 & \textbf{36.74} & \textbf{-112.40} & 110.10\\
& Wine & -82.31 & \textbf{5.19} & \textbf{-37.58} & 49.59 \\
& Breast cancer & -46.52 & \textbf{33.26} & \textbf{-27.0} & 81.47 \\
& House prices & -39.51 & \textbf{5.0} & \textbf{-38.12} & 9.54 \\
\hline\hline
\multirow{4}{*}{\rotatebox[origin=r]{90}{\twoLines{Decision}{tree}}}
& Iris &  -40.55 & \textbf{1.19} & \textbf{-0.73} & 4.06 \\
& Digits & -170.25 & \textbf{36.69} & \textbf{-110.48} & 114.78 \\
& Wine & -102.44 & \textbf{3.92} & \textbf{-34.38} & 66.92 \\
& Breast cancer & -43.44 & \textbf{0.01} & \textbf{-25.55} & 22.27 \\
& House prices & -40.49 & \textbf{0.01} & \textbf{-37.84} & 14.92 \\
\hline
\end{tabular}
\end{table}
We observe that our proposed density constraint consistently yields counterfactuals that have a higher density than the counterfactuals without any additional density/plausibility constraints - whereby we only observe a minor increase in computation time (e.g. from $30$ms to $70$ms per sample). However, the distance to the original sample is much higher for the "more plausible" counterfactuals than for arbitrary (e.g. closest) counterfactuals. This seems reasonable because one would expect that samples from a different class look quite differently.
In addition, we observe that the distances of the counterfactuals to the original samples on the Iris data set and Digits data set are more or less the same for both models, whereas the opposite is true for the wine, breast cancer and house prices data sets. This observation can be explained by the hypothesis that in the case of Iris and digits data set, both models learned a locally $\densitythreshold$-sufficient perfect classifier (Definition~\ref{def:locallyperfectclassifier}) at most samples - then Theorem~\ref{theorem:modelfreecounterfactuals:localversion} states that the counterfactuals are model independent which explains the observed numbers. Conversely, this suggests that the two classifiers learned on the other three data sets are quite different in the sense that they are not all locally $\densitythreshold$-sufficient perfect classifiers (Definition~\ref{def:locallyperfectclassifier}) at most samples - hence, the distances of the counterfactuals to the original samples are quite different.

\begin{figure}[!tb]
  \caption{Samples from the digit data set. \textit{First block:} Original samples. \textit{Second block:} Counterfactuals generated without any density/plausibility constraints. \textit{Third block:} Counterfactuals generated with our proposed density constraint. The corresponding labels are shown below each image - note that the shown labels of the counterfactuals are the requested labels.}
  \label{fig:digits:samples}

  \centering
  
  Original samples  
  
  \begin{minipage}[b]{0.19\textwidth}
    \includegraphics[width=\textwidth]{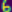}  
    \caption*{Label: 6} 
   \end{minipage}
  \hfill
  \begin{minipage}[b]{0.19\textwidth}
    \includegraphics[width=\textwidth]{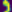}
    \caption*{Label: 9} 
  \end{minipage}
  \hfill
  \begin{minipage}[b]{0.19\textwidth}
    \includegraphics[width=\textwidth]{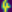}
    \caption*{Label: 4} 
  \end{minipage}
  \hfill  
  \begin{minipage}[b]{0.19\textwidth}
    \includegraphics[width=\textwidth]{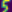}
    \caption*{Label: 5} 
  \end{minipage}
  \hfill  
  \begin{minipage}[b]{0.19\textwidth}
    \includegraphics[width=\textwidth]{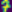}
    \caption*{Label: 7} 
  \end{minipage}  
 
  \rule[2ex]{12.2cm}{2.0pt}  
  Closest \textit{counterfactuals} under a \underline{softmax regression model}
  \vfill
  \begin{minipage}[b]{0.19\textwidth}
    \includegraphics[width=\textwidth]{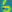}
    \caption*{Label: 3} 
  \end{minipage}
  \hfill
  \begin{minipage}[b]{0.19\textwidth}
    \includegraphics[width=\textwidth]{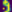}
    \caption*{Label: 0} 
  \end{minipage}
  \hfill
  \begin{minipage}[b]{0.19\textwidth}
    \includegraphics[width=\textwidth]{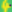}
    \caption*{Label: 6} 
  \end{minipage}
  \hfill
  \begin{minipage}[b]{0.19\textwidth}
    \includegraphics[width=\textwidth]{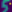}
    \caption*{Label: 7} 
  \end{minipage}
  \hfill
  \begin{minipage}[b]{0.19\textwidth}
    \includegraphics[width=\textwidth]{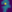}
    \caption*{Label: 1} 
  \end{minipage}
    
  Closest \textit{counterfactuals} under a \underline{decision tree model }
  \vfill
  \begin{minipage}[b]{0.19\textwidth}
    \includegraphics[width=\textwidth]{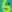}
    \caption*{Label: 3} 
  \end{minipage}
  \hfill
  \begin{minipage}[b]{0.19\textwidth}
    \includegraphics[width=\textwidth]{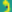}
    \caption*{Label: 0} 
  \end{minipage}
  \hfill
  \begin{minipage}[b]{0.19\textwidth}
    \includegraphics[width=\textwidth]{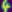}
    \caption*{Label: 6} 
  \end{minipage}
  \hfill
  \begin{minipage}[b]{0.19\textwidth}
    \includegraphics[width=\textwidth]{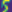}
    \caption*{Label: 7} 
  \end{minipage}
  \hfill
  \begin{minipage}[b]{0.19\textwidth}
    \includegraphics[width=\textwidth]{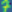}
    \caption*{Label: 1} 
  \end{minipage}
  
  \rule[2ex]{12.2cm}{2.0pt}    
  Closest \textit{plausible counterfactuals} under a \underline{softmax regression model}  
   \vfill
  \begin{minipage}[b]{0.19\textwidth}
    \includegraphics[width=\textwidth]{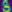}
    \caption*{Label: 3} 
  \end{minipage}
  \hfill
  \begin{minipage}[b]{0.19\textwidth}
    \includegraphics[width=\textwidth]{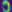}
    \caption*{Label: 0} 
  \end{minipage}
  \hfill
  \begin{minipage}[b]{0.19\textwidth}
    \includegraphics[width=\textwidth]{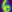}
    \caption*{Label: 6} 
  \end{minipage}
  \hfill
  \begin{minipage}[b]{0.19\textwidth}
    \includegraphics[width=\textwidth]{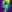}
    \caption*{Label: 7} 
  \end{minipage}
  \hfill
  \begin{minipage}[b]{0.19\textwidth}
    \includegraphics[width=\textwidth]{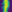}
    \caption*{Label: 1}
  \end{minipage}
  
  Closest \textit{plausible counterfactuals} under a \underline{decision tree model}
  \vfill
  \begin{minipage}[b]{0.19\textwidth}
    \includegraphics[width=\textwidth]{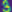}
    \caption*{Label: 3} 
  \end{minipage}
  \hfill
  \begin{minipage}[b]{0.19\textwidth}
    \includegraphics[width=\textwidth]{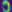}
    \caption*{Label: 0} 
  \end{minipage}
  \hfill
  \begin{minipage}[b]{0.19\textwidth}
    \includegraphics[width=\textwidth]{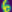}
    \caption*{Label: 6} 
  \end{minipage}
  \hfill
  \begin{minipage}[b]{0.19\textwidth}
    \includegraphics[width=\textwidth]{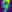}
    \caption*{Label: 7} 
  \end{minipage}
  \hfill
  \begin{minipage}[b]{0.19\textwidth}
    \includegraphics[width=\textwidth]{res/digits_origcfEX_logreg4.pdf}
    \caption*{Label: 1}
  \end{minipage}
\end{figure}

Furthermore, Fig.~\ref{fig:digits:samples} shows some samples from the digit data set and compares the counterfactuals generated with and without density constraints of both models.
Most of the samples in the second block - counterfactuals without any density/plausibility constraints - look like adversarials in the sense that the original label can be still recognized but the requested label can not be inferred. However, most of the samples in the third block - counterfactuals that have been computed with our proposed density constraint - look like samples from the requested target class. This suggests that our method in fact yields plausible counterfactuals. We also observe that the two models yield different counterfactuals in the second block but more or less exactly the same counterfactuals in the third block. As already discussed in the case of the very similar distances in Table~\ref{tab:experiments:results}, this can be explained by assuming that both models are (close to) locally $\densitythreshold$-sufficient perfect (Definition~\ref{def:locallyperfectclassifier}) at most samples, which confirms the observations as it is predicted by Theorem~\ref{theorem:modelfreecounterfactuals:localversion}.
However, please note that a visual inspection of some samples does not replace a proper evaluation by doing an expert user study and subsequent hypotheses testings.

\section{Discussion and Conclusion}
In this work, we proposed and studied a formal definition of plausible counterfactual explanations. In this definition we proposed to add density constraints to the optimization problem for computing counterfactual explanations to ensure that the resulting counterfactual is plausible in the given data domain. For practical purposes, we proposed convex approximations of a Gaussian mixture model to get tractable density constraints. These constraints give rise to convex optimization problems for computing plausible counterfactual explanations many common models like linear models and decision trees. In addition, these constraints allow to specify a lower bound on the density of the resulting counterfactual that is guaranteed to be full filled. Finally, we empirically evaluate our proposed methods on several data sets and observe that our method consistently yields counterfactual explanations that are located in high density regions. A visual inspection of samples from the digits data set suggests that in fact our method seems to yield plausible counterfactuals.

As future work, we plan to conduct a proper user study where humans judge the plausibility of generated counterfactual explanations - counterfactuals generated with and without density constrains. Furthermore, we want to explore density estimators for high dimensional data so that our method can be used for high dimensional data, too. We also plan to investigate how to add density constraints for computing counterfactual explanations of more complex models - in particular non-linear models (e.g. Deep neural networks). Lastly, our source code will be released as part of our open-source toolbox CEML~\cite{ceml}, a Python toolbox for computing counterfactual explanations of ML models, so that our proposed method can be easily used by practitioners.

\ifthenelse{\boolean{arxiv}}{
\appendix
\section{Proofs and Derivations}\label{sec:appendix}
\begin{enumerate}
\item
\begin{proof}[Theorem~\ref{theorem:modelfreecounterfactuals}]
For a given generating process $\generatingprocess$, zero risk classifiers exist iff the class-dependent densities are non-overlapping:
\begin{equation}
\begin{split}
&\exists\, \classifier: \underset{\x, \y \sim \density}{\E}[\Indicator\left(\classifier(\x) \neq \y\right)] = 0\\
&\Leftrightarrow \forall\,\x\in\setx:\;\density(\x, \y) \neq 0 \Leftrightarrow \density(\x, \y_i) = 0 \;\forall\,y_i\in\sety\setminus\{\y\}
\end{split}
\end{equation}
Therefore, for a zero risk classifier $\classifier$ it holds that:
\begin{equation}
\density(\x,\y) > 0 \implies \classifier(\x)=\y
\end{equation}
If follows that $\forall\,(\x,\ycf)\in\setx\times\sety$:
\begin{equation}
\begin{split}
&\underset{\xcf \,\in\,\setx}{\arg\min}\;\regularization(\xcf, \x) \quad \text{s.t. } \classifier(\xcf) = \ycf \; \land \; \density(\xcf,\ycf) \geq \densitythreshold\\
&\Leftrightarrow \underset{\xcf \,\in\,\setx}{\arg\min}\;\regularization(\xcf, \x) \quad \text{s.t. } \density(\xcf,\ycf) \geq \densitythreshold
\end{split}
\end{equation}
Thus, the constraint $\classifier(\xcf)=\ycf$ in Eq.~\eqref{eq:deltaplausiblecounterfactual:optprob} becomes redundant - the counterfactuals of zero risk classifiers do not depend on these classifiers.
\qed
\end{proof}

\item
\begin{proof}[Theorem~\ref{theorem:modelfreecounterfactuals:localversion}]
For a locally $\densitythreshold$-sufficient perfect classifier $\classifier$ (Definition~\ref{def:locallyperfectclassifier}) at $\x\in\setx$, it holds that:
\begin{equation}
\classifier(\x_{*}) = \ycf \;\;\forall\, \ycf \in \sety\setminus\{\y\},\; \x_{*}= \underset{\z\,\in\,\setx_{\densitythreshold}(\ycf)}{\arg\min}\regularization(\z, \x)
\end{equation}
where $\x_{*}$ is a $\densitythreshold$-plausible counterfactual (Definition~\ref{def:plausiblecounterfactual}) of $\x$.
It follows that:
\begin{equation}
\begin{split}
&\underset{\xcf \,\in\,\setx}{\arg\min}\;\regularization(\xcf, \x) \quad \text{s.t. } \classifier(\xcf) = \ycf \; \land \; \density(\xcf,\ycf) \geq \densitythreshold\\
&\Leftrightarrow \underset{\xcf \,\in\,\setx}{\arg\min}\;\regularization(\xcf, \x) \quad \text{s.t. } \density(\xcf,\ycf) \geq \densitythreshold
\end{split}
\end{equation}
Thus, the constraint $\classifier(\xcf)=\ycf$ in Eq.~\eqref{eq:deltaplausiblecounterfactual:optprob} becomes redundant - the counterfactuals of a sample $\x\in\setx$ of classifiers that are locally $\densitythreshold$-sufficient perfect at $\x$ do not depend on these classifiers.
\qed
\end{proof}

\item
\begin{proof}[Bound in Eq.~\eqref{eq:gmm:approx:bound}]
It holds that:
\begin{equation}\label{eq:gmm:component:nonnegative}
\pi_j \N(\x \mid \vec{\mu}_j,\mat{\Sigma}_j) \geq 0 \quad \forall\,j\in\{1,\dots,m\}
\end{equation}
Therefore, it follows that:
\begin{equation}
\densityestimator(\x) = \underset{j}{\max}\Big(\pi_j \N(\x \mid \vec{\mu}_j,\mat{\Sigma}_j)\Big) \leq \sum_{j=1}^m \pi_j \N(\x \mid \vec{\mu}_j,\mat{\Sigma}_j) = \densityestimatorgmm(\x)
\end{equation}
which proves the lower bound in Eq.~\eqref{eq:gmm:approx:bound}.

It holds that:
\begin{equation}\label{eq:gmm:maxcomponent}
\densityestimator(\x) = \underset{j}{\max}\Big(\pi_j \N(\x \mid \vec{\mu}_j,\mat{\Sigma}_j)\Big) \geq \pi_i \N(\x \mid \vec{\mu}_i,\mat{\Sigma}_i) \quad \forall\,i\in\{1,\dots,m\} 
\end{equation}
Because of Eq.~\eqref{eq:gmm:component:nonnegative} and Eq.~\eqref{eq:gmm:maxcomponent}, it follows that:
\begin{equation}
\densityestimatorgmm(\x) = \sum_{j=1}^m \pi_j \N(\x \mid \vec{\mu}_j,\mat{\Sigma}_j) \leq m \cdot \underset{j}{\max}\Big(\pi_j \N(\x \mid \vec{\mu}_j,\mat{\Sigma}_j)\Big) = m \cdot \densityestimator(\x)
\end{equation}
which proves the upper bound in Eq.~\eqref{eq:gmm:approx:bound}.
\qed
\end{proof}

\item
Eq.~\eqref{eq:gmm:approx:component:constraint} can be rewritten as the convex quadratic constraint Eq.~\eqref{eq:gmm:approx:component:finalconstraint}:
\begin{equation}
\begin{split}
& \pi_j\N(\x \mid \vec{\mu}_j,\mat{\Sigma}_j) \geq \densitythreshold \\
& \Leftrightarrow \log\Big(\pi_j\N(\x \mid \vec{\mu}_j,\mat{\Sigma}_j)\Big) \geq \log(\densitythreshold) \\
& \Leftrightarrow -\log\Big(\pi_j\N(\x \mid \vec{\mu}_j,\mat{\Sigma}_j)\Big) \leq -\log(\densitythreshold) \\
& \Leftrightarrow (\x - \vec{\mu}_j)^\top\mat{\Sigma}_j^{-1}(\x - \vec{\mu}_j) - 2\log(\pi_j) + d\log\left(2\pi\right) - \log\Big(\det(\mat{\Sigma}_j^{-1})\Big) \leq -2 \log(\densitythreshold)
\end{split}
\end{equation}
\end{enumerate}
}{}

%
%
%
\bibliographystyle{splncs04}
\bibliography{bibliography}
\end{document}